\documentclass[runningheads]{llncs}
\usepackage{graphicx}
\usepackage{amsmath}
\usepackage[ruled]{algorithm2e}
\usepackage{subfigure}
\usepackage{float}
\usepackage{hyperref}
\begin{document}
\title{Learning to gesticulate by observation using a deep generative
  approach\thanks{This work has been partially supported by the Basque
    Government (IT900-16 and Elkartek 2018/00114) and the Spanish
    Ministry of Economy and Competitiveness MINECO/FEDER (RTI
    2018-093337-B-100, MINECO/FEDER,EU). We gratefully acknowledge the
    support of NVIDIA Corporation with the donation of the Titan Xp
    GPU used for this research.}}
%
%
\author{Unai Zabala\inst{1}\orcidID{0000-0002-8196-2388} \and
Igor Rodriguez\inst{1}\orcidID{0000-0002-1432-102X} \and
Jos\'e Mar\'{\i}a Mart\'{\i}nez-Otzeta\inst{1}\orcidID{0000-0001-5015-1315}
\and 
Elena Lazkano\inst{1}\orcidID{0000-0002-7653-6210}
}
\authorrunning{U. Zabala et al.}
%
\institute{Computer Science and Artificial Intelligence\\ Faculty of Informatics, UPV/EHU\\ Manuel Lardizabal 1, 20018 Donostia\\
\email{igor.rodriguez@ehu.eus}\\
\url{http://www.sc.ehu.es/ccwrobot} }
\maketitle              
\begin{abstract}
  The goal of the system presented in this paper is to develop a
  natural talking gesture generation behavior for a humanoid robot, by
  feeding a Generative Adversarial Network (GAN) with human talking
  gestures recorded by a Kinect. A direct kinematic approach is used
  to translate from human poses to robot joint positions. The provided
  videos show that the robot is able to use a wide variety of
  gestures, offering a non-dreary, natural expression level.
\end{abstract}
\section{Introduction}
\label{sec:intro}

Social robotics \cite{breazeal04designing} aims to provide robots with
artificial social intelligence to improve human-machine interaction
and to introduce them in complex human contexts. The demand for
sophisticated robot behaviors requires to model and implement
human-like capabilities to sense, to process, and to act/interact
naturally by taking into account emotions, intentions, motivations,
and other related cognitive functions.

Talking involves spontaneous gesticulation; postures and movements are
relevant for social interactions even if they are subjective and
culture dependent. Aiming at building trust and making people feel
confident when interacting with them, socially acting humanoid robots
should show human-like talking gesticulation. Therefore, they need a
mechanism that generates movements that resembles humans' in terms of
naturalness. A previous work~\cite{rodriguez2016singing} made use of
gestures selected from a set of movements previously compiled. Those
gestures were then randomly concatenated and reproduced according to
the duration of the speech. That approach was prone to produce
repetitive movements and resulted in unnatural jerky
expression. 

The goal of the system presented in this paper is to develop a natural
talking gesture generation behavior for a humanoid robot. At this step
we aim to give a step forward by training a Generative Adversarial
Network (GAN) gesture generation system with movements captured
directly from humans. A Kinect sensor is used to track the skeleton of
the talking person and a GAN is trained to generate a richer and more
natural talking
gesticulation. 

Gestures (head, hands and arms movements) are used both to reinforce
the meaning of the words and to express feelings through non-verbal
signs. Among the different types of conversational movement of arms
and hands synchronised with the flow of the speech, beats are those
not associated with particular
meaning~\cite{mcneill1992hand}. References to talking gestures of the
present work will be limited to beats.

The robotic platform employed in the performed experiments is a
Softbank Robotics Pepper
robot~\footnote{\url{https://www.ald.softbankrobotics.com/en/robots/pepper}}. Currently,
our robot is controlled using the
\emph{naoqi\_driver}\footnote{\url{http://wiki.ros.org/naoqi\_driver}}
package that wraps the needed parts of
NAOqi~\footnote{\url{http://doc.aldebaran.com/2-5/naoqi/index.html}}
API and makes them available in
ROS\footnote{\url{http://www.ros.org}}.


\section{Related work}
\label{sec:related-work}
According to Beck et al.~\cite{beck2017body}, there are three main
robot motion generation approaches: manually creating motion, motion
capturing, and motion planning; for manual creation, it is required to
set each joint position of the humanoid robot for each key frame (time
step); the motion capture-based approach tries to mimic human
gestures, recording human movements and mapping these data to a
humanoid robot~\cite{penna2013whole}; and motion planning approach
relies on kinematics and/or dynamics equations to solve a geometric
task. They found that the motion capturing approach produces the most
realistic results, because the robot reproduces previously captured
human movements.

Motion capturing and imitation is a challenge because humans and
robots have different kinematic and dynamic structures. Motion capture
(MoCap) is the process of recording motion data through any type of
sensor. Applications of MoCap systems range from animation,
bio-mechanics, medicine to sports science, entertainment,
robotics~\cite{okamoto2014toward}\cite{zhang2018real} or even study of
animal behavior~\cite{schubert2016automatic}. MoCap systems rely on
optical technologies, and can be marker-based
(e.g. Vicon\footnote{\url{https://www.vicon.com/}}) or markerless like
RGB-D cameras. While the former ones provide more accurate results,
the latter ones are less prone to produce gaps (missing values) that
need to be estimated~\cite{tits2018robust,mehta2017vnect}. Many
approaches make use of the Kinect sensors due to its
availability~\cite{alibeigi2017inverse,fadli2017human,mukherjee2015inverse}.

No matter the motion capture system being used, there is a need to
transfer human motion to the robot joints. This can be done by direct
kinematics, adapting captured human joint angles to the robot. Or
alternatively, inverse kinematics calculates the necessary joint
positions given a desired end effector's pose.

On the other hand, generative models are probabilistic models capable
of generating all the values for a phenomenon. Unlike discriminative
models, they are able to generate not only the target variables but
also the observable ones \cite{tanwani2018generative}. They are used
in machine learning to (implicitly or explicitly) acquire the
distribution of the data for generating new samples. There are many
types of generative models. For instance, Bayesian
Networks~(BNs)\cite{castillo1997learning}, Gaussian Mixture Models
(GMMs)~\cite{everitt1981finite} and Hidden Markov Models
(HMMs)~\cite{rabiner89tutorial} are well known probability density
estimators.


Focusing on generative models used for motion generation,
in~\cite{kwon2006using} the authors propose the combination of
Principal Component Analysis (PCA) \cite{wold1987principal} and HMMs
for encoding different movement primitives to generate humanoid
motion. Tanwani~\cite{tanwani2018generative} uses HSMM (Hidden
Semi-Markov Models) for learning robot manipulation skills from
humans. Regarding on social robotics, some generative approaches are
being applied with different objectives. In \cite{manfre2016automatic}
Manfr\`{e} et al. use HMMs for dance creation and in a later work they
try variational auto-encoders again for the same
purpose~\cite{augello2017creative}.

Deep learning techniques have also been applied to generative models,
giving rise to deep generative models. A taxonomy of such models can
be found in \cite{goodfellow2016nips}. In particular Generative
Adversarial Networks (GANs)~\cite{goodfellow2014generative} are
semi-supervised emerging models that basically learn how to generate
synthetic data from the given training data. GANs are deep generative
models capable to implicitly acquire the probability density function
in the training data, being able to automatically discover the
internal structure of datasets by learning multiple levels of
abstraction\cite{lecun2015deep}. Gupta et al. \cite{gupta2018social}
extend the use of GANs to generate socially acceptable motion
trajectories in crowded scenes in the scope of self-driving
cars. In~\cite{rodriguez2019spontaneous} GANs showed to overcome other
generative approaches such as HMM and GMM when confronted to the task
of motion generation. In that work, movements produced synthetically
(using choregraph) were used to train the different generative
approaches. Instead, in this paper we feed the GAN with movements
obtained by observing and capturing human talking gestures.

\section{Developed approach to enhance robot spontaneity}
\label{sec:approach}


The GAN used in the current approach takes as input only
proprioceptive joint position information. In order to feed the GAN
with natural motion data, a motion capturing approach is
employed. Thus, these two aspects are exhaustively described here on.


\subsection{Human motion capture and imitation}
\label{sec:skeleton-tracking}
In~\cite{rodriguez2014humanizing}, direct kinematics was used to
teleoperate a NAO robot. Human skeleton obtained with a Kinect was
tracked and arm movements were replicated by the robot, while walking
motions were commanded by using different spatial key movements. As
the goal was to teleoperate the robot, there was no need for subtle
and continuous motion since the arms only required to reach single
poses when demanded by the operator. On the contrary, gesticulation
requires continuous arm motion and involves also hands, head and
fingers. Although the present work makes use of a similar motion
capture and mapping system, the presented system has been enriched to
cover all the aspects involved.

The Kinect uses structured light (depth map) and machine learning to
infer body position~\cite{kinect}. The OpenNI based
\texttt{skeleton\_markers}\footnote{\url{http://wiki.ros.org/skeleton\_markers}}
ROS package is able to extract in real time the 15 joints associated
to the human
skeleton. 
\vspace{-0.5cm}
\subsubsection{Mapping human's arms into robot's space:} \hfill \break
\label{subsec:arm-control}

\begin{figure}[!htbp]
\centering
\includegraphics[width=0.9\columnwidth]{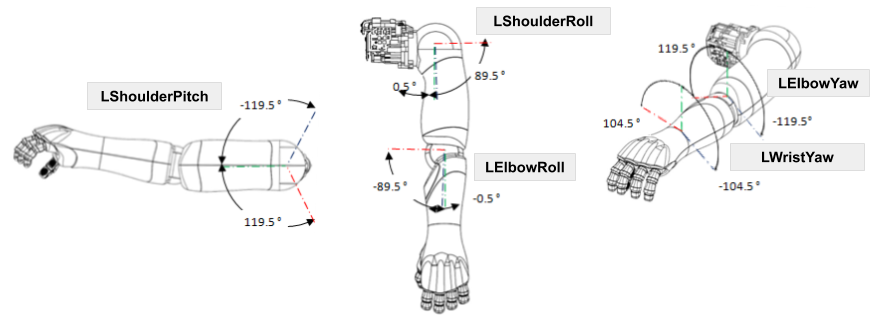}
\caption{Pepper's left arm joints and actuators (from Softbanks official Pepper's user guide\protect\ref{fntsoft}).}

\label{fig:left-arm-joints}
\end{figure}

Human arms have 7 degrees of freedom (DoF): a spherical joint at the
shoulder, a revolute joint at the elbow and a spherical joint at the
wrist. On the contrary, our humanoid's arms have 5 DoF: two at the
shoulder (pitch and yaw) and elbow (yaw and roll), and one at the
wrist (yaw) (fig.~\ref{fig:left-arm-joints}). Thereby, the movement
configurations of human and robot arms differ.

To transform the Cartesian coordinates obtained from the Kinect into
\textit{Pepper}'s coordinate space a joint control approach was
employed. Note that the transformations are performed to the reference
system of each individual joint, not to a robot's global reference
frame. On the following explanation we will focus on the left arm. The
analysis of the right arm is similar and it will be omitted
here.

\textit{Pepper}'s left arm has five
joints\footnote{\label{fntsoft}\url{http://doc.aldebaran.com/2-8/family/pepper\_technical/joints\_pep.html}}
(see fig.~\ref{fig:left-arm-joints}): shoulder roll ($LS_\alpha$) and
pitch ($LS_\beta$), elbow roll ($LE_\alpha$) and yaw ($LE_\gamma$) and
wrist yaw ($LW_\gamma$). The \texttt{skeleton\_markers} package can
not detect the operator's hands' yaw motion and thus, $LW_\gamma$
joints cannot be reproduced using the skeleton information. We chose
another approach for $LW_\gamma$, that will be explained later on.


In order to calculate the shoulder's roll angle ($LS_\alpha$) we use
the dot product of the distance vector between both shoulders
($\overline{LRS}$) and the length vector between the shoulder and the
elbow ($\overline{LSE}$). Note that, before computing that product,
$\overline{LRS}$ and $\overline{LSE}$ vectors must be normalized. The
$LS_\alpha$ angle is calculated in the Kinect's coordinate space,
therefore, it must be transformed into the robot's coordinate space by
rotating it $\frac{-\pi}{2}$ radians
(eq.~\ref{eq:alpha_shoulder_roll}).

\begin{equation}
\label{eq:alpha_shoulder_roll}
\begin{split}
LS_\alpha = \arccos{(\overline{LRS} \cdot \overline{LSE})}\\
LS^{robot}_\alpha = LS_\alpha - \frac{\pi}{2}
\end{split}
\end{equation}

Elbow's roll ($LE_\alpha$) angle is calculated in the same way as
shoulder's roll angle ($LS_\alpha$) but the length vector between the
shoulder and the elbow ($\overline{LSE}$) and the length vector
between the elbow and the hand ($\overline{LEH}$) are used
instead. Again, those vectors need to be normalized and transformed to
the robot's space, in this case by rotating it $-\pi$
radians.


With respect to elbow's yaw angle ($LE_\gamma$) calculation we use
only the y and z components of the $\overline{LEH}$ vector. After
normalizing $\overline{LEH}$, eq.~\ref{eq:elbow_yaw} is applied to
obtain the $LE_\gamma$. Lastly, a range conversion is needed to get
$LE^{robot}_\gamma$ (from [$\frac{\pi}{2}$, $\pi$] to
[$-\frac{\pi}{2}$, 0] and from [$-\pi$, $-\frac{\pi}{2}$] to [0,
$\pi$]).

\begin{equation}
\label{eq:elbow_yaw}
\begin{split}
LE_\gamma = \arctan{\frac{\overline{LEH}_z}{\overline{LEH}_y}}\\
LE^{robot}_\gamma = rangeConv(LE_\gamma)
\end{split}
\end{equation} 

To conclude with the joints, the shoulder pitch angle ($LS_\beta$) is
calculated by measuring the angle between the shoulder to elbow vector
and the $z$ axis.  $z=0$ occurs with the arm extended at $90^\circ$
with respect to the torso. Thus, lowering the arm produces negative
pitch angle while raising it above the shoulder produces positive
angular values.

The $LS_\beta$ can be defined as:

\begin{equation}
\label{eq:elbow pitch}
\begin{split}
\| A \| = LSE_z \mbox{\quad (by \quad definition)}\\
\sin{(LS_\beta)}=\frac{\| A \|}{\| \overline{LSE} \|}=\frac{\| A \|}{1} \\
LS^{robot}_\beta = \arcsin{(LSE_z)}
\end{split}
\end{equation}

where $LSE_z$ is the $Z$ coordinate of the shoulder to elbow vector.  

\subsubsection{Mapping human's hands into robot's space:} \hfill \break
\label{subsec:hands}

Hands movements are common in humans while talking. We do rotate wrist
and open and close hands, move fingers constantly. Unfortunately, the
skeleton capturing system we are using does not allow to detect such
movements. It is possible though to capture the state of the hands
using a different approach.

\begin{figure}[!htbp]
  \centering
  \includegraphics[width=0.6\columnwidth]{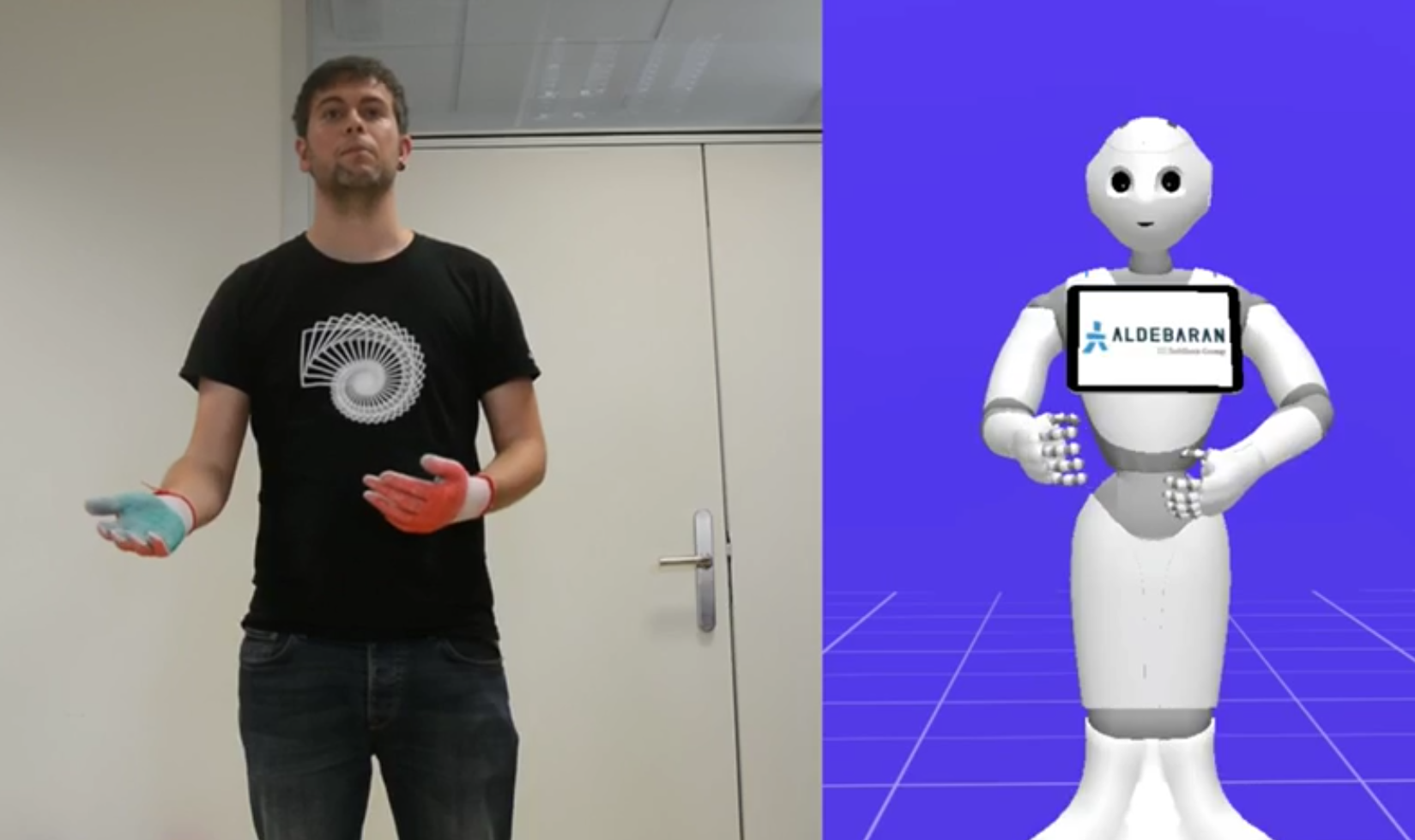}
  \caption{Snapshot of a data capture session.}
  \label{fig:igor-gloves}
\end{figure}

The developed solution forces the user to wear coloured gloves, green
in the palm of the hand and red in the back
(fig.~\ref{fig:igor-gloves}).  While the human talks, hands
coordinates are tracked and those positions are mapped into the image
space and a subimage is obtained for each hand. Angular information is
afterwards calculated by measuring the number of pixels ($max$) of the
outstanding color in a subimage. Eq.~\ref{eq:hands} shows the
procedure for the left hand. $N$ is a normalizing constant and
$maxW_\gamma$ stands for the maximum wrist yaw angle of the robot.

\begin{equation}
\left\{ 
\begin{array}{ll} LW^{robot}_\gamma =   max/N \times maxW_\gamma & \mbox{ if $max$ is palm}\\
LW^{robot}_\gamma =  \frac{max-N}{N} \times maxW_\gamma
  & \mbox{ otherwise}
  \end{array}\right.
\label{eq:hands}
\end{equation}
\vspace{-0.5cm}


In addition $LE_\gamma$ is modified when humans palms are up (subimage
has only green pixels) to easy the movement of the robot.

Regarding the fingers, as they cannot be tracked, their position is
randomly set at each skeleton frame.

\subsubsection{Mapping human's head into robot's space} 
\begin{figure}[!htbp]
\centering
\includegraphics[width=0.5\columnwidth]{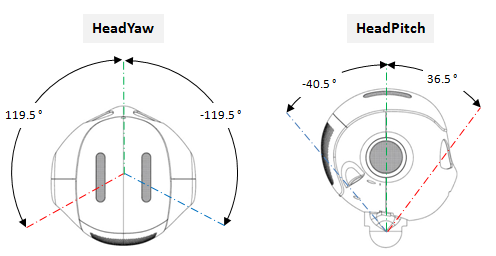}
\caption{Pepper's head joints and actuators (from Softbanks official Pepper's user guide\protect{\textsuperscript{\ref{fntsoft}}}).}
\label{fig:head-joints}
\end{figure}

\hfill \break
\label{sec:head-control}
Humans move the head while talking and thus, head motion should also
be captured and mapped.  The robot's head has 2 DoFs that allow the
head to move left to right (yaw) and up and down (pitch) as shown in
fig.~\ref{fig:head-joints}.


The Kinect skeleton tracking program gives us the (neck and) head 3D
positions. The approach taken for mapping the yaw angle to the robot's
head consists of applying a gain $K_1$ to the human's yaw value, once
transformed into the robot space by a $-\frac{\pi}{2}$ rotation
(eq.~\ref{eq:head-yaw}).

\begin{equation}
\label{eq:head-yaw}
H^{robot}_\gamma = K_1 \times H_\beta 
\end{equation}

In order to approximate head's pitch angle, the head to neck vector
($\overline{HN}$) is calculated and rotated $-\frac{\pi}{2}$ and then,
its angle is obtained (eq.~\ref{eq:head-pitch}). Note that robot's
head is an ellipsoid instead of an sphere. To avoid unwanted head
movements a lineal gain is applied to the final value.

\begin{equation}
\label{eq:head-pitch}
H^{robot}_\beta = \arctan{(rotate(\overline{HN}, -\frac{\pi}{2}))} + |K_2*H_\gamma|
\end{equation}

\subsection{Generative model}
\label{sec:gan}
GAN networks are composed by two different interconnected
networks. The \emph{Generator} (\textit{G}) network generates possible
candidates so that they are as similar as possible to the training
set. The second network, known as \emph{Discriminator} (\textit{D}),
judges the output of the first network to discriminate whether its
input data are ``real'', namely equal to the input data set, or if
they are ``fake'', that is, generated to trick with false
data. 

The training dataset given to the \textit{D} network contained 2018
unit of movements (UM), being each UM is a sequence of 4 consecutive
poses, and each pose 14 float numbers corresponding to joint values of
head, arms, wrists and hands (finger opening). These samples were
recorded by registering 5 different persons talking, about 9 minutes
overall.

\begin{figure}[!htbp]
\centering
\includegraphics[width=0.6\columnwidth]{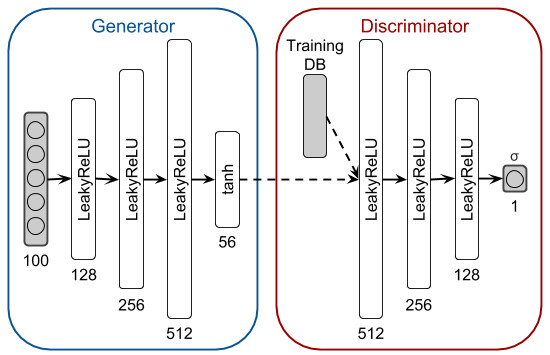}
\caption{GAN setup for talking gesture generation.}
\label{fig:gan-setup}
\end{figure}

The \textit{D} network is thus trained using that data to learn its
distribution space and its input dimension is 56. On the other hand,
the \textit{G} network is seeded through a random input with a uniform
distribution in the range [$-1, 1$] and with a dimension of 100. The
\textit{G} intends to produce as output gestures that belong to the
real data distribution and that the \textit{D} network would not be
able to correctly pick out as generated. Fig.~\ref{fig:gan-setup}
depicts the architecture the generator and discriminator networks.

GAN has been trained for 2000 epochs and its hyper-parameter have been
tuned experimentally; we set up a batch size of 16, a learning rate of
0.0002, Adam \cite{kingma2014adam} as optimization method, and
$\beta_1$ = 0.5 and $\beta_2$ = 0.999 as its parameters.

\section{Results}
\label{sec:results}
The obtained robot performance can be appreciated in the following videos:
\begin{enumerate}
\item The first
  video\footnote{\url{https://www.youtube.com/watch?v=iW1566ozbdg}}
  shows some instants recorded during the process of generating the
  database of movements captured through the motion capturing and
  imitation mechanisms. On the left side the participant talking and
  gesticulating is displayed, while the simulated robot mimicking the
  movements in real time (without GAN) is shown in the right side.


\item A second
  video\footnote{\url{https://www.youtube.com/watch?v=1It\_Y\_AEnts}}
  shows the evolution of the robot behavior during different steps of
  the training process. The final number of epochs was empirically set
  to 2000 for the model that has been integrated into the gesture
  generation system.
\end{enumerate}

Notice that the temporal length of the audio intended to be pronounced
by the robot determines the number of units of movement required to
the generative model. Thus, the execution of those units of movements,
one after the other, defines the whole movement displayed by the
robot.

\section{Conclusions and further work}
\label{sec:conclusions}
In this work a talking gesture generation system has been developed
using a GAN feeded with natural motion data obtained through a motion
capturing and imitation system. The suitability of the approach is
demonstrated with a real robot. Results show that the obtained robot
behavior is appropriate, and thanks to the movement variability the
robot expresses itself with naturalness.


As further work, we intend to improve the skeleton capture process by
using more robust systems, such as
OpenPose~\cite{cao2017realtime,cao2018openpose} or
wrnchAI\footnote{\url{https://wrnch.ai/}} that allow to capture more
detailed movements. In this way, the speakers would not need to wear
the gloves, that somehow are conditioning them. Moreover, speakers
tend to behave in an constricted way when recorded. A more powerful
skeleton tracker system would allow to use recorded videos from real
talks and to build a more objective database.  With respect to the
mapping process, in~\cite{mukherjee2015inverse} direct kinematics is
compared with two inverse kinematics approaches and the neuro fuzzy
approach seems to improve the direct one. The use of a more effective
method to translate human poses to robot poses could also produce
better movements.

The work presented here pretends to be the starting point to acquire a
richer gesture set, such as emotion-based gestures or context related
gestures. Moreover, a generator conditioned on the sentence/word
itself would correspond to how humans use their gestures to emphasize
their communication.




%
%
%
%

\bibliographystyle{splncs04}
\bibliography{srbiblio}

\begin{thebibliography}{10}
\providecommand{\url}[1]{\texttt{#1}}
\providecommand{\urlprefix}{URL }
\providecommand{\doi}[1]{https://doi.org/#1}

\bibitem{alibeigi2017inverse}
Alibeigi, M., Rabiee, S., Ahmadabadi, M.N.: Inverse kinematics based human
  mimicking system using skeletal tracking technology. Journal of Intelligent
  {\&} Robotic Systems  \textbf{85}(1),  27--45 (Jan 2017)

\bibitem{augello2017creative}
Augello, A., Cipolla, E., Infantino, I., Manfr{\`{e}}, A., Pilato, G., Vella,
  F.: Creative robot dance with variational encoder. CoRR
  \textbf{abs/1707.01489} (2017)

\bibitem{beck2017body}
Beck, A., Yumak, Z., Magnenat-Thalmann, N.: Body movements generation for
  virtual characters and social robots. In: Social signal processing, chap.~20,
  pp. 273--286. Cambridge University Press (2017)

\bibitem{breazeal04designing}
Breazeal, C.: Designing sociable robots. Intelligent Robotics and Autonomous
  Agents, MIT Press, Cambridge MA, USA (2004)

\bibitem{cao2018openpose}
Cao, Z., Hidalgo, G., Simon, T., Wei, S.E., Sheikh, Y.: Open{P}ose: realtime
  multi-person 2{D} pose estimation using {P}art {A}ffinity {F}ields. In: arXiv
  preprint arXiv:1812.08008 (2018)

\bibitem{cao2017realtime}
Cao, Z., Simon, T., Wei, S.E., Sheikh, Y.: Realtime multi-person {2D} pose
  estimation using part affinity fields. In: CVPR (2017)

\bibitem{castillo1997learning}
Enrique~Castillo, J.M.G., Hadi, A.S.: Learning Bayesian Networks. Expert
  Systems and Probabilistic Network Models. Monographs in computer science. New
  York: Springer-Verlag (1997)

\bibitem{everitt1981finite}
Everitt, B., Hand, D.: Finite mixture distributions. Chapman and Hall (1981)

\bibitem{fadli2017human}
Fadli, H., Machbub, C., Hidayat, E.: Human gesture imitation on {NAO} humanoid
  robot using {Kinect} based on inverse kinematics method. In: International
  Conference on Advanced Mechatronics, Intelligent Manufacture, and Industrial
  Automation {(ICAMIMIA)}. {IEEE} (2015)

\bibitem{goodfellow2016nips}
{Goodfellow}, I.: {NIPS Tutorial: Generative Adversarial Networks}. ArXiv
  e-prints  (Dec 2017)

\bibitem{goodfellow2014generative}
Goodfellow, I., Pouget-Abadie, J., Mirza, M., Xu, B., Warde-Farley, D., Ozair,
  S., Courville, A., Bengio, Y.: Generative adversarial nets. In: Advances in
  neural information processing systems. pp. 2672--2680 (2014)

\bibitem{gupta2018social}
Gupta, A., Johnson, J., Fei{-}Fei, L., Savarese, S., Alahi, A.: Social {GAN:}
  socially acceptable trajectories with {Generative Adversarial Networks}. CoRR
   \textbf{abs/1803.10892} (2018), \url{http://arxiv.org/abs/1803.10892}

\bibitem{kingma2014adam}
Kingma, D.P., Ba, J.: Adam: A method for stochastic optimization. arXiv
  preprint arXiv:1412.6980  (2014)

\bibitem{kwon2006using}
Kwon, J., Park, F.C.: {Using Hidden Markov Models to Generate Natural Humanoid
  Movement}. In: International Conference on Intelligent Robots and Systems
  (IROS). {IEEE/RSJ} (2006)

\bibitem{lecun2015deep}
LeCun, Y., Bengio, Y., Hinton, G.: Deep learning. Nature  \textbf{521}(7553),
  436--444 (2015)

\bibitem{kinect}
MacCormick, J.: How does the {Kinect} work?
  http://pages.cs.wisc.edu/~ahmad/kinect.pdf (accessed June 3, 2019)

\bibitem{manfre2016automatic}
Manfr{\`e}, A., Infantino, I., Vella, F., Gaglio, S.: An automatic system for
  humanoid dance creation. Biologically Inspired Cognitive Architectures
  \textbf{15}, ~1--9 (2016)

\bibitem{mcneill1992hand}
McNeill, D.: Hand and mind: What gestures reveal about thought. University of
  Chicago press (1992)

\bibitem{mehta2017vnect}
Mehta, D., Sridhar, S., Sotnychenko, O., Rhodin, H., Shafiei, M., Seidel, H.P.,
  Xu, W., Casas, D., Theobalt, C.: {VNect: Real-time 3D Human Pose Estimation
  with a Single RGB Camera}. ACM Trans. Graph.  \textbf{36}(4),  44:1--44:14
  (Jul 2017)

\bibitem{mukherjee2015inverse}
Mukherjee, S., Paramkusam, D., Dwivedy, S.K.: Inverse kinematics of a {NAO}
  humanoid robot using {Kinect} to track and imitate human motion. In:
  International Conference on Robotics, Automation, Control and Embedded
  Systems {(RACE)}. {IEEE} (2015)

\bibitem{okamoto2014toward}
Okamoto, T., Shiratori, T., Kudoh, S., Nakaoka, S., Ikeuchi, K.: Toward a
  dancing robot with listening capability: Keypose-based integration of lower-,
  middle-, and upper-body motions for varying music tempos. IEEE Transactions
  on Robotics  \textbf{30},  771--778 (Jun 2014).
  \doi{10.1109/TRO.2014.2300212}

\bibitem{penna2013whole}
Poubel, L.P.: Whole-body Online Human Motion Imitation by a Humanoid Robot
  Using Task Specification. Master's thesis, Ecole Centrale de Nantes--Warsaw
  University of Technology (2013)

\bibitem{rabiner89tutorial}
Rabiner, L.R.: A tutorial on {Hidden Markov Models} and selected applications
  in speech recognition. In: Proceedings of the {IEEE}. vol.~77, pp. 257--286
  (1989)

\bibitem{rodriguez2016singing}
Rodriguez, I., Astigarraga, A., Ruiz, T., Lazkano, E.: Singing minstrel robots,
  a means for improving social behaviors. In: IEEE International Conference on
  Robotics and Automation (ICRA). pp. 2902--2907 (2016)

\bibitem{rodriguez2014humanizing}
Rodriguez, I., Astigarraga, A., Jauregi, E., Ruiz, T., Lazkano, E.: Humanizing
  {NAO} robot teleoperation using {ROS}. In: International Conference on
  Humanoid Robots (Humanoids) (2014)

\bibitem{rodriguez2019spontaneous}
Rodriguez, I., Mart\'\i{}nez-Otzeta, J.M., Irigoien, I., Lazkano, E.:
  Spontaneous talking gestures using generative adversarial networks. Robotics
  and Autonomous Systems  \textbf{114},  57 -- 65 (2019)

\bibitem{schubert2016automatic}
Schubert, T., Eggensperger, K., Gkogkidis, A., Hutter, F., Ball, T., Burgard,
  W.: Automatic bone parameter estimation for skeleton tracking in optical
  motion capture. In: International Conference on Robotics and Automation
  {(ICRA)}. {IEEE} (2016)

\bibitem{tanwani2018generative}
Tanwani, A.K.: Generative Models for learning Robot Manipulation. Ph.D. thesis,
  {\'E}cole Polytechnique F{\'e}d{\'e}ral de Laussane (EPFL) (2018)

\bibitem{tits2018robust}
Tits, M., Tilmanne, J., Dutoit, T.: Robust and automatic motion-capture data
  recovery using soft skeleton constraints and model averaging. PLOS ONE
  \textbf{13}(7),  1--21 (Jul 2018)

\bibitem{wold1987principal}
Wold, S., Esbensen, K., Geladi, P.: Principal component analysis. Chemometrics
  and intelligent laboratory systems  \textbf{2}(1-3),  37--52 (1987)

\bibitem{zhang2018real}
Zhang, Z., Niu, Y., Yan, Z., Lin, S.: Real-time whole-body imitation by
  humanoid robots and task-oriented teleoperation using an analytical mapping
  method and quantitative evaluation. Applied Sciences  \textbf{8}(10) (2018),
  \url{https://www.mdpi.com/2076-3417/8/10/2005}

\end{thebibliography}

\end{document}